\pdfoutput=1

\documentclass[11pt]{article}

\usepackage[]{_preamble/acl}

\usepackage{times}
\usepackage{latexsym}

\usepackage[T1]{fontenc}

\usepackage[utf8]{inputenc}

\usepackage{microtype}
\usepackage{subfigure}

\usepackage{amsmath}
\usepackage{amsfonts}
\usepackage{placeins}

\usepackage{_preamble/default_packages}
\usepackage{_preamble/colab}
\usepackage{_preamble/mjb}
\usepackage{_preamble/local_defs}

\title{Word2Box: Capturing Set-Theoretic Semantics of Words using Box Embeddings}

\author{Shib Sankar Dasgupta\thanks{*Equal Contributions.} , \ Michael Boratko\footnotemark[1] , \ Siddhartha Mishra , \\ \bf{Shriya Atmakuri} , \ \bf{Dhruvesh Patel} , \ \bf{Xiang Lorraine Li} , \ \bf{Andrew McCallum} \\
        Manning College of Information \& Computer Sciences \\
        University of Massachusetts Amherst\\
    \texttt{\{ssdasgupta,mboratkro,siddharthami,satmakuri,}\\ 
    \texttt{dhruveshpate,xiangl,mccallum\}@cs.umass.edu}}

\begin{document}
\maketitle
\begin{abstract}

Learning representations of words in a continuous space is perhaps the most fundamental task in NLP, however words interact in ways much richer than vector dot product similarity can provide. 
Many relationships between words can be expressed set-theoretically, for example adjective-noun compounds (eg. "red cars"$\subseteq$"cars") and homographs (eg. "tongue"$\cap$"body" should be similar to "mouth", while "tongue"$\cap$"language" should be similar to "dialect") have natural set-theoretic interpretations.
Box embeddings are a novel region-based representation which provide the capability to perform these set-theoretic operations.
In this work, we provide a fuzzy-set interpretation of box embeddings, and learn box representations of words using a set-theoretic training objective.
We demonstrate improved performance on various word similarity tasks, particularly on less common words, and perform a quantitative and qualitative analysis exploring the additional unique expressivity provided by \wb.
\end{abstract}
\section{Introduction}
The concept of learning a distributed representation for a word has fundamentally changed the field of natural language processing.
The introduction of efficient methods for training vector representations of words in Word2Vec \cite{word2vec}, and later GloVe \cite{glove} as well as FastText \cite{fasttext} revolutionized the field, paving the way for the recent wave of deep architectures for language modeling, all of which implicitly rely on this fundamental notion that a word can be effectively represented by a vector.

While now ubiquitous, the concept of representing a word as a single point in space is not particularly natural. All senses and contexts, levels of abstraction, variants and modifications which the word may represent are forced to be captured by the specification of a single location in Euclidean space. It is thus unsurprising that a number of alternatives have been proposed.

\begin{figure}[t]
\includegraphics[width=\linewidth]{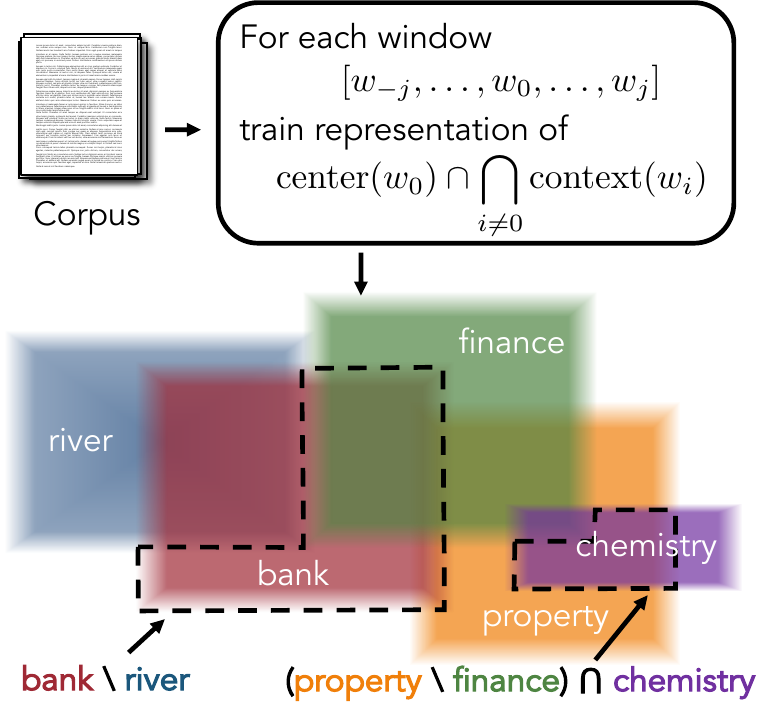}
\caption{
\label{fig:training_process}
Given a corpus, Gumbel Boxes are trained as a fuzzy sets representing sets of windows with given center or context words. The representations can then be queried using multiple set-theoretic operations. In the graphic, \texttt{bank}$\setminus$\texttt{river} overlaps highly with \texttt{finance}, and we would also expect high overlap with other boxes (not depicted) such as \texttt{firm} or \texttt{brokerage} (see \Cref{table:set_difference}). Similarly, we would expect boxes for chemical properties such as \texttt{hardness} or \texttt{solubility} to overlap with the dotted region indicating \texttt{property}$\setminus$\texttt{finance}$\cap$\texttt{chemistry}, and indeed we do observe such overlaps in the \wb model (see \Cref{table:set_difference_intersection}).}
\end{figure}

Gaussian embeddings \cite{vilnis2015word} propose modeling words using densities in latent space as a way to explicitly capture uncertainty. Poincaré embeddings \cite{poincare_glove} attempt to capture a latent hierarchical graph between words by embedding words as vectors in hyperbolic space. Trained over large corpora via similar unsupervised objectives as vector baselines, these models demonstrate an improvement on word similarity tasks, giving evidence to the notion that vectors are not capturing all relevant structure from their unsupervised training objective.

A more recent line of work explores region-based embeddings, which use geometric objects such as disks \citep{suzuki2019hyperbolic}, cones \citep{order_embedding, poe, hyperbolic_cone}, and boxes \citep{hard_box} to represent entities. These models are often motivated by the need to express asymmetry, benefit from particular inductive biases, or benefit from calibrated probabilistic semantics. In the context of word representation, their ability to represent words using geometric objects with well-defined intersection, union, and difference operations is of interest, as we may expect these operations to translate to the words being represented in a meaningful way.


In this work, we introduce \wb, a region-based embedding for words where each word is represented by an $n$-dimensional hyperrectangle or ``box''. Of the region-based embeddings, boxes were chosen as the operations of intersection, union, and difference are easily calculable.
Specifically, we use a variant of box embeddings known as Gumbel boxes, introduced in \cite{gumbel_box}.
Our objective (both for training and inference) is inherently set-theoretic, not probabilistic, and as such we first provide a fuzzy-set interpretation of Gumbel boxes yielding rigorously defined mathematical operations for intersection, union, and difference of Gumbel boxes.

We train boxes on a large corpus in an unsupervised manner with a continuous bag of words (CBOW) training objective,

using the intersection of boxes representing the context words as the representation for the context. The resulting model demonstrates improved performance compared to vector baselines on a large number of word similarity benchmarks.
We also compare the models' abilities to handle set-theoretic queries, and find that the box model outperforms the vector model 90\% of the time.
Inspecting the model outputs qualitatively also demonstrates that \wb can provide sensible answers to a wide range of set-theoretic queries.

\section{Background}
\label{sec:background}

\paragraph{Notation} Let $V=\{v_i\}_{i=1}^N$ denote the vocabulary, indexed in a fixed but arbitrary order. A sentence $\mathbf s = (s_1, \ldots, s_j)$ is simply a (variable-length) sequence of elements in our vocab $s_i \in V$.
We view our corpus $C = \{\mathbf s_i\}$ as a multiset
\footnote{A \defterm{multiset} is a set which allows for repetition, or equivalently a sequence where order is ignored.}
of all sentences in our corpus.
Given some fixed "window size" $\ell$, for each word $s_i$ in a sentence $\mathbf s$ we can consider the window centered at $i$,
\[\mathbf w_i = [s_{i-\ell}, \ldots, s_i, \ldots, s_{i + \ell}],\]
where we omit any indices exceeding the bounds of the sentence. Given a window $\mathbf w_i$ we denote the center word using $\cen(\mathbf w_i) = s_i$, and denote all remaining words as the context $\con(\mathbf w_i)$. We let $C_W$ be the multiset of all windows in the corpus. 



\subsection{Fuzzy sets}
\label{sec:fuzzy sets}

Given any ambient space $U$ a set $S\subseteq U$ can be represented by its characteristic function $\ind_S: U\to \{0, 1\}$ such that $\ind_S(u)=1 \iff u\in S$. This definition can be generalized to consider functions $m: U \to [0, 1]$, in which case we call the pair $A=(U, m)$ a \defterm{fuzzy set} and $m = m_A$ is known as the \defterm{membership function} \citep{zadeh1965fuzzy,fuzzy-zadeh-1996}.
There is historical precedent for the use of fuzzy sets in computational linguistics \cite{zhelezniak2018dont,note-on-fuzzy-language}. More generally, fuzzy sets are naturally required any time we would like to learn a set representation in a gradient-based model, as hard membership assignments would not allow for gradient flow.

In order to extend the notion of set intersection to fuzzy sets, it is necessary to define a \defterm{t-norm}, which is a binary operation $\top:[0,1]\times [0,1]\to[0,1]$ which is commutative, monotonic, associative, and equal to the identity when either input is 1. The $\min$ and product operations are common examples of t-norms. Given any t-norm, the intersection of fuzzy sets $A$ and $B$ has membership function $m_{A \cap B}(x) = \top(m_A(x), m_B(x))$. Any t-norm has a corresponding t-conorm which is given by $\bot(a,b) = 1-\top(1-a, 1-b)$; for $\min$ the t-conorm is $\max$, and for product the t-conorm is the probabilistic sum, $\bot_\text{sum}(a,b) = a+b-ab$. This defines the union between fuzzy sets, where $m_{A \cup B}(x) = \bot(m_A(x), m_B(x))$. Finally, the complement of a fuzzy set simply has member function $m_{A^c}(x) = 1 - m_A(x)$.

\subsection{Box embeddings}
\label{sec:box embeddings}

Box embeddings, introduced in \cite{hard_box}, represent elements $\mathbf x$ of some set $X$ as a Cartesian product of intervals,
\begin{align}
\label{eq: hard box definition}
\begin{split}
    \Box(\mathbf x) &\defeq \prod_{i=1}^d [x_i^\boxmin, x_i^\boxmax]\\
    &= [x_1^\boxmin, x_1^\boxmax]\times \cdots \times[x_d^\boxmin, x_d^\boxmax] \subseteq \RR^d.
\end{split}
\end{align}
The volume of a box is simply the multiplication of the side-lengths,
\begin{equation*}
    |\Box(\mathbf x)| = \prod_{i=1}^d \max(0, x_i^\boxmax - x_i^\boxmin),
\end{equation*}
and when two boxes intersect, their intersection is
\begin{multline*}
    \Box(\mathbf x)\cap \Box(\mathbf y) \\=
    \prod_{i=1}^d
    [\max(x_i^\boxmin, y_i^\boxmin), \min(x_i^\boxmax, y_i^\boxmax)].
\end{multline*}
Boxes are trained via gradient descent, and these hard $\min$ and $\max$ operations result in large areas of the parameter space with no gradient signal. \citet{gumbel_box} address this problem by modeling the corners of the boxes $\{x_i^\pm\}$ with Gumbel random variables, $\{X_i^\pm\}$, where the probability of any point $\mathbf z \in \RR^d$ being inside the box $\Box_G(\mathbf x)$ is given by
\[P(\mathbf z \in \GBox(\mathbf x)) = \prod_{i=1}^d P(z_i > X_i^-)P(z_i < X_i^+).\]
For clarity, we will denote the original ("hard") boxes as $\Box$, and the Gumbel boxes as $\GBox$. The Gumbel distribution was chosen as it was min/max stable, thus the intersection $\GBox(\mathbf x) \cap \GBox(\mathbf y)$ which was defined as a new box with corners modeled by the random variables $\{Z_i^\pm\}$ where
\begin{equation*}
    Z_i^- \defeq \max(X_i^-, Y_i^-) \text{ and } 
    Z_i^+ \defeq \min(X_i^+, Y_i^+)
\end{equation*}
is actually a Gumbel box as well. 

\citet{boratko2021min} observed that
\begin{multline}
\label{eq: gumbel product}
P(\mathbf z \in \GBox(\mathbf x) \cap \GBox(\mathbf y)) =\\ P(\mathbf z \in \GBox(\mathbf x)) P(\mathbf z \in \GBox(\mathbf y)),
\end{multline}
and also provided a rigorous probabilistic interpretation for Gumbel boxes when embedded in a space of finite measure, leading to natural notions of "union" and "intersection" based on these operations of the random variables \cite{boratko2021min}.

In this work, we do not embed the boxes in a space of finite measure, but instead interpret them as \emph{fuzzy sets}, where the above probability (of a point $\mathbf z$ being inside the Gumbel box) acts as a soft membership function.

\section{Fuzzy Sets of Windows}
\label{sec: Fuzzy Sets of Windows}
In this section, we describe the motivation for using fuzzy sets to represent words, starting with an approach using traditional sets.

First, given a word $v \in V$, we can consider the windows centered at $v$,
\[\CenWin(v) \defeq\{w \in C_W : \cen(w) = v\},\]
and the set of windows whose context contains $v$,
\[\ConWin(v) \defeq\{w \in C_W : \con(w) \ni v\}.\]
Note that $\cen_W$ is a function which takes in a word and returns a set of windows, whereas $\cen$ is a function which takes in a window and returns the center word, and a similar distinction holds for $\con_W$ and $\con$.

A given window is thus contained inside the intersection of the sets described above, namely
\begin{multline*}
[w_{-j}, \ldots, w_0, \ldots, w_{j}] \\\in \CenWin(w_0) \cap \bigcap_{i \ne 0} \ConWin(w_i).
\end{multline*}
As an example, the window 
\[\mathbf w = \text{"quick brown fox jumps over"},\]
is contained inside the $\CenWin(\text{"fox"})$ set, as well as $\ConWin(\text{"quick"})$, $\ConWin(\text{"brown"})$, $\ConWin(\text{"jumps"})$, $\ConWin(\text{"over"})$.
With this formulation, the intersection of the $\ConWin$ sets provide a natural choice of representation for the context. We might hope that $\CenWin(v)$ provides a reasonable representation for the word $v$ itself, however by our set theoretic definition for any $u \ne v$ we have $\CenWin(u) \cap \CenWin(v) = \emptyset$.


We would like the representation of $u$ to overlap with $v$ if $u$ has "similar meaning" to $v$, 
i.e. we would like to consider
\[\CenSim(v) \defeq\{w \in W : \cen(w) \text{ similar to } v\}.\]
A crisp definition of \emph{meaning} or \emph{similarity} is not possible \citep{simlex999, ws353} due to individual subjectivity. Inter-annotator agreement for \citet{simlex999} is only 0.67, for example, which makes it clear that $\CenSim(v)$ could not possibly be represented as a traditional set. Instead, it seems natural to consider $\CenSim(v)$ as represented by a fuzzy set $(W, m)$, where $m(w)\in [0,1]$ can be thought of as capturing graded similarity between $v$ and $\cen(w)$.\footnote{For an even more tangible definition, we can consider $m(w)$ the percentage of people who consider $u$ to be similar to $\cen(w)$ when used in context $\con(w)$.} In the same way, we can define
\[\ConSim(v) \defeq \{ w \in W : \con(v)\ni w \text{ similar to } v\},\]
which would also be represented as a fuzzy set.\footnote{Note that this gives a principled reason to use different representation for $\widetilde{\cen_W}(v)$ and $\widetilde{\con_W}(v)$, as they fundamentally represent different sets.}

As we wish to capture these similarities with a machine learning model, we now must find trainable representations of fuzzy sets.

\begin{remark}
Our objective of learning trainable representations for these sets provides an additional practical motivation for using fuzzy sets - namely, the hard assignment of elements to a set is not differentiable. Any gradient-descent based learning algorithm which seeks to represent sets will have to consider a smoothed variant of the characteristic function, which thus leads to fuzzy sets.
\end{remark}

\section{Gumbel Boxes as Fuzzy Sets}
\label{sec: Gumbel Boxes as Fuzzy Sets}
In this section we will describe how we model fuzzy sets using Gumbel boxes \cite{gumbel_box}. As noted in \Cref{sec:box embeddings}, the Gumbel Box model represents entities $\mathbf x \in X$ by $\GBox(\mathbf x)$ with corners modeled by Gumbel random variables $\{X_i^\pm\}$. The probability of a point $\mathbf z \in \RR^d$ being inside this box is
\[\resizebox{\linewidth}{!}{$\displaystyle P(\mathbf z \in \GBox(\mathbf x)) = \prod_{i=1}^d P(z_i > X_i^-)P(z_i < X_i^+).$}\]
Since this is contained in $[0,1]$, we have that $(\RR^d, P(\mathbf z \in \GBox(\mathbf x))$ is a fuzzy set. For clarity, we will refer to this fuzzy set as $\FBox(\mathbf x)$.

The set complement operation has a very natural interpretation in this setting, as $\FBox(\mathbf x)^c$ has membership function $1 - P(\mathbf z \in \GBox(\mathbf x))$, that is, the probability of $\mathbf z$ not being inside the Gumbel box. The product t-norm is a very natural choice as well, as the intersection $\FBox(\mathbf x) \cap \FBox(\mathbf y)$ will have membership function $P(\mathbf z \in \GBox(\mathbf x))P(\mathbf z \in \GBox(\mathbf y))$,
which is precisely the membership function associated with $\GBox(\mathbf x) \cap \GBox(\mathbf y)$, where here the intersection is between Gumbel boxes as defined in \citet{gumbel_box}.
Finally, we find that the membership function for the union $\FBox(\mathbf x) \cup \FBox(\mathbf y)$ is given (via the t-conorm) by
\begin{multline}
\label{eq: union membership function}
P(\mathbf z \in \GBox(\mathbf x)) + P(\mathbf z \in \GBox(\mathbf y)) - \\P(\mathbf z \in \GBox(\mathbf x) P(\mathbf z \in \GBox(\mathbf y)).
\end{multline}

\begin{remark}
Prior work on Gumbel boxes had not defined a union operation on Gumbel boxes, however \eqref{eq: union membership function} has several pleasing properties apart from being a natural consequence of using the product t-norm. First, it can be directly interpreted as the probability of $\mathbf z$ being inside $\GBox(\mathbf x)$ or $\GBox(\mathbf y)$. Second, if the Gumbel boxes were embedded in a space of finite measure, as in \citet{boratko2021min}, integrating \eqref{eq: union membership function} would yield the probability corresponding to $P(\Box(\mathbf x) \cup \Box(\mathbf y))$.
\end{remark}





To calculate the size of the fuzzy set $\FBox(\mathbf x)$ we integrate the membership function over $\RR^d$,
\[|\FBox(\mathbf x)| = \int_{\RR^d} P(\mathbf z \in \GBox(\mathbf x)) \, d\mathbf z.\]
The connection between this integral and that which was approximated in \citet{gumbel_box} is provided by Lemma 3 of \citet{boratko2021min}, and thus we have
\[\resizebox{\linewidth}{!}{$\displaystyle|\FBox(\mathbf x)|\approx \prod_{i=1}^d\beta \log\left(1 + \exp\left(\frac{\mu_i^+ - \mu_i^-}\beta - 2 \gamma\right)\right)$}\]
where $\mu_i^-, \mu_i^+$ are the location parameters for the Gumbel random variables $X_i^-, X_i^+$, respectively.
As mentioned in \Cref{sec:box embeddings}, Gumbel boxes are closed under intersection, i.e. $\GBox(\mathbf x) \cap \GBox(\mathbf y)$ is also a Gumbel box, which implies that the size of the fuzzy intersection
\begin{align*}
|&\FBox(\mathbf x) \cap \FBox(\mathbf y)| \\
&=\int_{\RR^d} P(\mathbf z \in \GBox(\mathbf x))P(\mathbf z \in \GBox(\mathbf y))\, d\mathbf z\\
&=\int_{\RR^d} P(\mathbf z \in \GBox(\mathbf x)\cap\GBox(\mathbf y))\, d\mathbf z
\end{align*}
can be approximated as well. As both of these are tractable, integrating \eqref{eq: union membership function} is also possible via linearity. Similarly, we can calculate the size of fuzzy set differences, such as
\begin{multline*}
|\FBox(\mathbf x) \setminus \FBox(\mathbf y)|=\\
\int_{\RR^d} P(\mathbf z \in \GBox(\mathbf x))[1 - P(\mathbf z \in \GBox(\mathbf y))]\, d\mathbf z.
\end{multline*}
By exploiting linearity and closure under intersection, it is possible to calculate the size of arbitrary fuzzy intersections, unions, and set differences, as well as any combination of such operations.
\begin{remark}
If our boxes were embedded in a space of finite measure, as in \citet{boratko2021min}, the sizes of these fuzzy sets would correspond to the intersection, union, and negation of the binary random variables they represent.
\end{remark}
\section{Training}
\label{sec: Training}
In this section we describe our method of training fuzzy box representations of words, which we refer to as \wb.

In \Cref{sec: Fuzzy Sets of Windows} we defined the fuzzy sets $\CenSim(v)$ and $\CenSim(v)$, and in \Cref{sec: Gumbel Boxes as Fuzzy Sets} we established that Gumbel boxes can be interpreted as fuzzy sets, thus for \wb we propose to learn center and context box representations
\begin{align*}
\CenBox(v) &\defeq \FBox(\CenSim(v))\\
\ConBox(v) &\defeq \FBox(\ConSim(v)).
\end{align*}

Given a window,
$\mathbf w = [w_{-j}, \ldots, w_0, \ldots, w_j]$, we noted that $\mathbf w$ must exist in the intersection,
\begin{equation}
    \CenSim(w_0) \cap \bigcap_{i\ne 0} \ConSim(w_i)
\end{equation}
and thus we consider a max-margin training objective where the score for a given window is given as
\begin{equation}
    f(\mathbf w) \defeq \bigg |\CenBox(w_0) \cap \bigcap_{i \ne 0} \ConBox(w_i) \bigg|.
\end{equation}
To create a negative example $\mathbf w'$ we follow the same procedure as CBOW from \citet{word2vec}, replacing center words with a word sampled from the unigram distribution raised to the $3/4$. We also subsample the context words as in \citet{word2vec}. As a vector baseline, we compare with a \wv model trained in CBOW-style. We attach the source code with supplementary material.
\section{Experiments and Results}
\label{sec:exp-results}
We evaluate both \wv and \wb on several quantitative and qualitative tasks that cover the aspects of semantic similarity, relatedness, lexical ambiguity, and uncertainty. Following the previous relevant works \cite{gaussian_density,lexical-ambiguity, baroni-2012}, we train on the lemmatized WaCkypedia corpora \cite{wackypedia}, specifically ukWaC which is an English language corpus created by web crawling. After additional pre-processing (details in \cref{app: preprocessing}) the corpus contains around 0.9 billion tokens, with just more than 112k unique tokens in the vocabulary. Noting that an $n$-dimensional box actually has $2n$ parameters (for min and max coordinates), we compare 128-dimensional \wv embeddings and 64-dimensional \wb embeddings for all our experiments. We train over 60 different models for both the methods for 10 epochs using random sampling on a wide range of hyperparameters (please refer to \cref{app:training_hyperparam} for details including learning rate, batch size, negative sampling, sub-sampling threshold, etc.).
In order to ensure that the only difference between the models was the representation itself, we implemented a version of \wv in PyTorch, including the negative sampling and subsampling procedures recommended in \cite{word2vec}, using the original implementation as a reference. As we intended to train on GPU, however, our implementation differs from the original in that we use Stochastic Gradient Descent with varying batch sizes. We provide our source code at \url{https://github.com/iesl/word2box}.

\subsection{Word Similarity Benchmarks}
We primarily evaluate our method on several word similarity benchmarks: SimLex-999 \citep{simlex999}, WS-353 \citep{ws353}, YP-130 \citep{YP}, MEN \citep{men},  MC-30 \citep{MC}, RG-65 \citep{rg}, VERB-143 \citep{verb-143}, Stanford RW \citep{rw}, Mturk-287 \citep{Mturk-287} and Mturk-771 \citep{Mturk-771}. These datasets consist of pairs of words (both noun and verb pairs) that are annotated by human evaluators for semantic similarity and relatedness.

In \cref{tab:similarity_results} we compare the \wb and \wv models which perform best on the similarity benchmarks.
We observe that \wb outperforms \wv (as well as the results reported by other baselines) in the majority of the word similarity tasks. We outperform \wv by a large margin in Stanford RW and YP-130, which are the rare-word datasets for noun and verb respectively. Noticing this effect, we 
enumerated the frequency distribution of each dataset. 
The datasets fall in different sections of the frequency spectrum, e.g., Stanford RW \cite{rw} only contains rare words which make its median frequency to be 5,683, whereas WS-353 (Rel) \cite{ws353} contains many more common words, with a median frequency of 64,490.
We also observe a larger relative performance improvement over \wv on other datasets which have low to median frequency words, e.g. MC-30, MEN-Tr-3K, and RG-65, all with median frequency less than 25k.
The order they appear in the table and the subsequent plots is lowest to highest frequency, left to right.
Please refer to Appendix B for details.

\begin{figure}
    \includegraphics[width=0.5\textwidth]{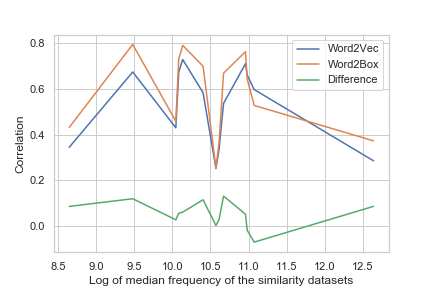}
    \caption{\small This plot depicts the gain in correlation score for \wb against \wv is much higher for the low and mid frequency range.}
\label{fig:freqvscorr}
\end{figure}

\begin{figure*}[ht!]
        \includegraphics[width=.25\textwidth]{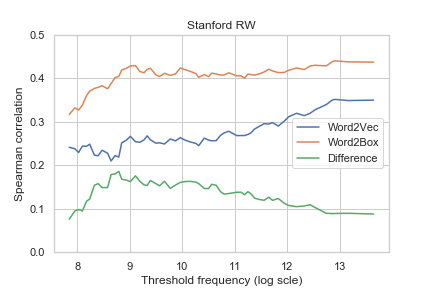}\hfill
        \includegraphics[width=.25\textwidth]{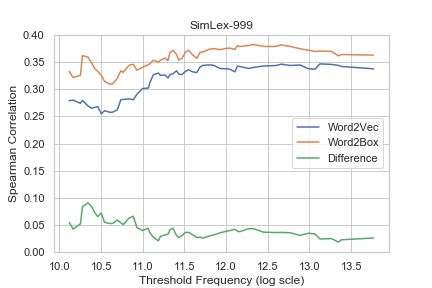}\hfill
        \includegraphics[width=.25\textwidth]{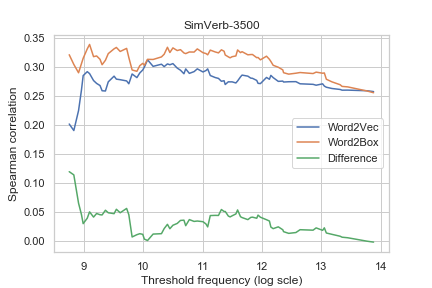}\hfill
        \includegraphics[width=.25\textwidth]{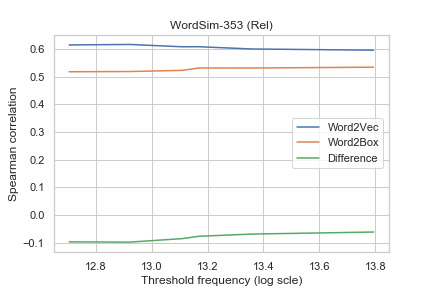}
        \caption{\small We plot the Spearman's correlation score vs Threshold frequency in log scale for Stanford RW, Simelex-999
        SimVerb-3500, WS-353 (Rel). The correlation value is calculated on the word pairs where both of them have frequency less than the threshold frequency.
        }
    \label{fig:freqvscorr2}
    \end{figure*}

In figure \ref{fig:freqvscorr}, we see that \wb outperforms \wv more significantly with less common words. In order to investigate further, we selected four datasets (RW-Stanford (rare words), Simelex-999, SimVerb-3500,WS-353 (Rel)), truncated them at a frequency threshold, and calculated the correlation for different levels of this threshold. In figure \ref{fig:freqvscorr2}, we demonstrate how the performance gap between \wb and \wv changes as increasing amount frequent words are added to these similarity datasets. We posit that the geometry of box embeddings is more flexible in the way it handles sets of mutually disjoint words  (such as rare words) which all co-occur with a more common word. Boxes have exponentially many corners, relative to their dimension, allowing extreme flexibility in the possible arrangements of intersection to represent complicated co-occurrances.

\begin{table*}
\resizebox{\linewidth}{!}{%

\begin{tabular}{lrrrrrrrrrrrrr}
\toprule
{} &  Stanford RW &  RG-65 &  YP-130 &  MEN &  MC-30 &  Mturk-287 &  SimVerb-3500 &  SimLex-999 &  Mturk-771 &  WS-353 (Sim) &  WS-353 (All) &  WS-353 (Rel) &  VERB-143 \\
\midrule
*Poincaré & --- & 75.97 & --- & --- & 80.46 & --- & 18.90 & 31.81 & --- & --- & 62.34 & --- & ---\\
*Gaussian & --- & 71.00 & 41.50 & 71.31 & 70.41 & --- & --- & 32.23 & --- & 76.15 & 65.49 & 58.96 & ---\\
\midrule
\wv  &        40.25 &  66.80 &   43.77 & 68.45 &  75.57 &      61.83 &         23.58 &       37.30 &      59.90 &              75.81 &              \textbf{69.01} &              \textbf{61.29} &     31.97 \\
\wb &        \textbf{45.08} &  \textbf{81.45} &   \textbf{51.6} & \textbf{73.68} &  \textbf{87.12} &      \textbf{70.62} &         \textbf{29.71} &       \textbf{38.19} &      \textbf{68.51} &              \textbf{78.60} &              \textbf{68.68} &              60.34 &     \textbf{48.03} \\
\bottomrule
\bottomrule
\end{tabular}%
}
    \caption{\small Similarity: We evaluate our box embedding model \wb against a standard vector baseline \wv. For comparison, we also include the reported results for Gaussian and Poincaré embeddings, however we note that these may not be directly comparable as many other aspects (eg. corpus, vocab size, sampling method, training process, etc.) may be different between these models.}
    \label{tab:similarity_results}
\end{table*}

\subsection{Set Theoretic Operations}
All the senses, contexts and abstractions of a word can not be captured accurately using a point vector, and must be captured with sets. In this section, we evaluate our models capability of representing sets by performing set operations with the trained models.

\subsubsection{Dataset}
Homographs, words with identical spelling but distinct meanings, and polysemous words are ideal choice of probe for this purpose, as demonstrated by the "bank", "river" and "finance" example of \Cref{fig:training_process}. We constructed set-theoretic logical operations on words based on common polysemous words and homographs \cite{homographs}. For example, the word "property" will have association with words related both "asset" and "attribute", and thus the union of the later two should be close to the original word "property". Likewise, the intersection set of "property" and "math" should contain many words related to mathematical properties or concepts.

To this end, we created a dataset consisting of triples $(A, B, C)$ where $A \circ B$ should yield a set similar to $C$, for various set-theoretic operations $\circ$. In this task, given two words $A$ and $B$ and a set theoretic operation $\circ$, we try to find the rank of word $C$ in the sorted list based on the set similarity (vector similarity scores for the vectors) score between $A \circ B$ and all words in the vocab. The dataset consists of 52 examples for both Union and Negation, and 20 examples for Intersection. The details of the dataset can be found in \Cref{app:set_theoretic_queries}.

\subsubsection{Quantitative Results}
 In \Cref{tab:set_operations}, we report the percentage of times \wb outperforms \wv , i.e. the model yields better rank for the word $C$. Note that it is not clear how to design the union, difference or the intersection operations with vectors. We consider several relevant choices, including component-wise operations (addition, subtraction, min and max) which yield a representation for $A \circ B$, as well as operations which operate on the scores - eg. score max pooling ranks each word $X$ using $\max(A \cdot X, B \cdot X)$, and similarly for score min pooling.
The purpose of these operations is to mimic the essence of union and intersection in the vector space, however, it is evident that the trained vector geometry is not harmonious to this construction as well.

We observe that almost of all the values are more than 0.9, meaning that \wb yields a higher rank for the target $C$ than \wv over 90\% of the time. This empirically validates that our model is indeed capturing the underlying set theoretic aspects of the words in the corpus. 

\begin{table}
\begin{tabular}{|l|lll}
\hline
\diagbox[width=8em,leftsep=0em,rightsep=0em]{\wv}{\wb}                  & \multicolumn{1}{l|}{$A \cap B$} & \multicolumn{1}{l|}{$A \setminus B$} & \multicolumn{1}{l|}{$A \cup B$} \\ \hline
$(A + B) \cdot X$          & 0.90                            & 0.92                                & 0.98                                           \\ \cline{1-1}
$(A - B) \cdot X$ & 0.90                            & 0.65                                & 0.80                                           \\ \cline{1-1}
$\max(A, B) \cdot X$       & 0.95                            & 0.86                                & 0.86                                           \\ \cline{1-1}
$\min(A, B) \cdot X$       & 0.90                            & 0.75                                & 0.92                                           \\ \cline{1-1}
$\max(A \cdot X, B \cdot X)$ & 0.95                            & 0.84                                & 0.94                                           \\ \cline{1-1}
$\min(A \cdot X, B \cdot X)$ & 1.0                             & 0.80                                & 0.84                                           \\ \cline{1-1}
\end{tabular}
    \caption{
    \label{tab:set_operations}
    \small Percentage of queries for which \wb set operations return the target word with higher rank than the given vector operation for \wv. Scores higher than $0.5$ means that \wb outperformed \wv. For subsequent qualitative comparisons we take the vector operation which performs most favorably for \wv.
    }
\end{table}

\subsubsection{Qualitative Analysis}
In this section, we present some interesting examples of set theoretic queries on words, with different degrees of complexities.  For all the tables in this section, we perform the set-operations on the query words and present the ranked list of most similar words to the output query. Many of these queries are based on the aforementioned homographs, for which there are natural expectations of what various set-theoretic operations should capture. Our results are presented in \Cref{table:set_operations_comparison}-\ref{table:set_difference_intersection}.

The results in \Cref{table:similarity} look reasonable for both models, as is to be expected since this is simply the similarity function for each model. Even increasing to a single intersection, as in \Cref{table:intersection}, starts to demonstrate that \wv may often return very low-frequency words.
In \Cref{table:set_difference}, we observe that set difference of "property" and "land" yields a set of words that are related to attributes of science subjects, eg. algebra or chemistry. We wanted to examine how the model would handle more complicated queries, for example if we first perform "property"$\setminus$"finance" and then further intersect with "algebra" or "chemistry", does the introduction of the relatively high-frequency "finance" term cause the model to struggle to recapture these items? In \Cref{table:set_difference_intersection} we observe that the outputs for \wb do indeed correspond to properties of those sub-fields of science, whereas the results in \wv focus strongly on "finance". In general, we observe better consistency of \wb with all the example logical queries.

\begin{table*}
    \resizebox{\linewidth}{!}{%
    \centering
\begin{tabular}{lllllllll}
\toprule
\multicolumn{3}{|c|}{\wb} & \multicolumn{6}{c|}{\wv}\\
\hline
$($bank$ \cap $river$) \cap X$ & $($bank$ \cup $river$) \cap X$ & $($bank$ \setminus $river$) \cap X$ & $($bank$ + $river$) \cdot X$ & $($bank$ - $river$) \cdot X$ & $\max($bank$, $river$) \cdot X$ & $\max($bank$, $river$) \cdot X$ & $\max($bank$ \cdot X, $river$ \cdot X)$ & $\min($bank$ \cdot X, $river$ \cdot X)$ \\
\midrule
         headwaters &           tributary &                 barclays &       tributaries &           cheques &            tributary &               vipava &                  tributaries &                       gauley \\
          tributary &              valley &                     hsbc &         tributary &        tymoshenko &          tributaries &              quabbin &                   headwaters &                   pymatuning \\
               lake &          headwaters &                  banking &        headwaters &       receivables &                 prut &               irwell &                    tributary &                 'utricularia \\
              basin &           reservoir &                citigroup &           nakdong &          citibank &              chambal &         trabajadores &                    headwater &                      luangwa \\
            estuary &               gorge &                 citibank &            vipava &          eurozone &           headwaters &        chattahoochee &               distributaries &                       vipava \\
              creek &                lake &                     firm &           estuary &            brinks &               larrys &          tributaries &                       larrys &                 guadalquivir \\
             valley &                 dam &                      ipo &            larrys &         defrauded &              nakdong &               belait &                        kobuk &                         suir \\
          reservoir &           headwater &                brokerage &         headwater &        courtaulds &             waterway &           bougouriba &                      estuary &                    meenachil \\
              canal &            junction &                interbank &      distributary &         refinance &            loyalsock &                canal &                       ijssel &                    tributary \\
         floodplain &               creek &                     kpmg &           luangwa &          mortgage &          'hyperolius &               glomma &                 distributary &                      battuta \\
\bottomrule
\end{tabular}
}
    \caption{Output of \wb and \wv for various set operations}
    \label{table:set_operations_comparison}
\end{table*}

\begin{table*}
\resizebox{\linewidth}{!}{%
    \centering
\begin{tabular}{l>{\raggedright}p{0.55\linewidth}>{\raggedright\arraybackslash}p{0.55\linewidth}}
\toprule
{} &                                                     \wb  &                                                                                                              \wv \\
A              &    $A \cap X$                                                                   &  $A \cdot X$                                                                                                                               \\
\midrule
bank           &                capital settlement airline hotel gateway treasury firm government loan casino &                                      debit depositors securities kaupthing interbank subprime counterparty citibank fdic nasdaq \\
economics      &            education architecture politics economy literature faculty agriculture phd journalism  &  microeconomic keynesian microeconomics minored macroeconomics econometrics sociology thermodynamics evolutionism structuralist \\
microeconomics &  economics mathematics physics philosophy theory technology economist principle research analysis &                  microeconomic initio germline instantiation zachman macroeconomics oxoglutarate glycemic noncommutative pubmed \\
property       &                   land register status manor purpose locality premise landmark site residence &                           easement infringes burgage krajobrazowy chattels policyholder leasehold intestate liabilities ceteris \\
rock           &                              music pop mountain cave band blues dance groove hot disco &                                           shoegaze rhyolitic punk britpop mafic outcrops metalcore bluesy sedimentary quartzite \\
\bottomrule
\end{tabular}
}
\caption{Similarity outputs for \wb and \wv}
\label{table:similarity}
\end{table*}
\begin{table*}
\centering
\resizebox{\linewidth}{!}{%
\begin{tabular}{ll>{\raggedright}p{0.5\linewidth}>{\raggedright\arraybackslash}p{0.5\linewidth}}
\toprule
      &          &                                                                   \wb  &                                                                      \wv \\
A & B &                                                                 $(A \cap B) \cap X$                             &  $(A + B) \cdot X$                                                                                             \\
\midrule
girl & boy &                           kid girls schoolgirl teenager woman boys child baby teenage orphan &                    shoeshine nanoha soulja schoolgirl yeller beastie jeezy crudup 'girl rahne \\
property & burial &                cemetery bury estate grave interment tomb dwelling site gravesite sarcophagus &        interment moated interred dunams ceteris burials catafalque easement deeded inhumation \\
      & historical &  historic estate artifact archaeological preserve ownership patrimony heritage landmark site &  krajobrazowy burgage easement kravis dilapidation tohono intangible domesday moated laertius \\
      & house &               estate mansion manor residence houses tenement building premise buildings site &       leasehold mansion tenements outbuildings estate burgage bedrooms moated burgesses manor \\
tongue & body &                                       eye mouth ear limb lip forehead anus neck finger penis &      tubercle ribcage meatus diverticulum forelegs radula tuberosity elastin foramen nostrils \\
      & language &            dialect idiom pronunciation meaning cognate word accent colloquial speaking speak &         fluently dialects vowels patois languages loanwords phonology lingala tigrinya fluent \\
\bottomrule
\end{tabular}
    }
\caption{Comparison of set intersection operation}
\label{table:intersection}
\end{table*}

\begin{table*}
\resizebox{\linewidth}{!}{%
\begin{tabular}{ll>{\raggedright}p{0.50\linewidth}>{\raggedright\arraybackslash}p{0.50\linewidth}}
\toprule
         &      &                                                                               \wb  &                                                                                                   \wv  \\
A & B &                                                             $(A \setminus B) \cap X$                                                  & $(A - B) \cdot X$                                                                                                                           \\
\midrule
algebra & finance &    homomorphism isomorphism automorphism abelian algebraic bilinear topological morphism spinor homeomorphism &  homeomorphic unital homomorphisms nilpotent algebraically projective holomorphic propositional nondegenerate endomorphism \\
bank & finance &                             wensum junction neman mouth tributary downstream corner embankment forks sandwich &                                         shaddai takla thrombus gauley paria epenthetic chibchan urubamba foremast bolshaya \\
         & river &                                    barclays hsbc banking citigroup citibank firm ipo brokerage interbank kpmg &                            cheques tymoshenko receivables citibank eurozone brinks defrauded courtaulds refinance mortgage \\
chemistry & finance &            biochemistry superconductor physics physic eutectic heat isotope fluorescence yttrium spectroscopy &                  augite alkyne desorption phosphorylating dimorphism fumarate hypertrophic empedocles hydratase enantiomer \\
property & land &  homotopy isomorphism involution register bijection symplectic eigenvalue idempotent compactification lattice &                                                brst stieltjes l'p repressor absurdum doesn conjugates nonempty didn wouldn \\
\bottomrule
\end{tabular}
}
\caption{Comparison of set difference operation}
\label{table:set_difference}
\end{table*}
\begin{table*}
\resizebox{\linewidth}{!}{%
\begin{tabular}{lll>{\raggedright}p{0.42\linewidth}>{\raggedright\arraybackslash}p{0.42\linewidth}}
\toprule
         &         &           &                                                                   \wb  &                                                                          \wv  \\
A & B & C &                                                                                     $((A \setminus B) \cap C) \cap X$                      &                                                                                     $(A - B + C) \cdot X$                  \\
\midrule
property & finance & algebra &  laplacian nilpotent antiderivative lattice surjective automorphism invertible homotopy integer integrand &    expropriate extort refco underwrite reimburse refinance parmalat refinancing brokerage privatizing \\
         &         & chemistry &    eutectic desiccant allotrope phenocryst hardness solubility monoclinic hygroscopic nepheline trehalose &  refinance brokerage burgage stockbroking refinancing warranties reimburse madoff privatizing valorem \\
\bottomrule
\end{tabular}
}
\caption{Comparison of set difference followed by intersection operation}
\label{table:set_difference_intersection}
\end{table*}

\section{Related Work}

Learning distributional vector representations from a raw corpus was introduced in \citet{word2vec}, quickly followed by various improvements \citep{glove, fasttext}. More recently, vector representations which incorporate contextual information have shown significant improvements \cite{elmo, bert, gpt2, gpt3}. As these models require context, however, Word2Vec-style approaches are still relevant in settings where such context is unavailable.

Hyperbolic representations \cite{poincare, hyperbolic_cone, hyperbolic_graph} have become popular in recent years. Most related to our setting, \citet{poincare_glove} propose a hyperbolic analog to GloVe, with the motivation that the hyperbolic embeddings will discover a latent hierarchical structure between words.\footnote{Reported results are included in \cref{tab:similarity_results} as "Poincaré"}
\citet{vilnis2015word} use Gaussian distributions to represent each word, and KL Divergence as a score function. \footnote{Reported results are included in \cref{tab:similarity_results} as "Gaussian"} \citet{gaussian_density} extended such representations by adding certain thresholds for each distribution. For a different purpose, \citet{ren2020beta} use Beta Distributions to model logical operations between words. Our work can be seen as a region-based analog to these models.

Of the region-based embeddings, \citet{suzuki2019hyperbolic} uses hyperbolic disks, and \citet{hyperbolic_cone} uses hyperbolic cones, however these are not closed under intersection nor are their intersections easily computable. \citet{order_embedding} and  \citet{poe} use an axis-aligned cone to represent a specific relation between words/sentences, for example an entailment relation. \citet{hard_box} extends \citet{poe} by adding an upper-bound, provably increasing the representational capacity of the model.
\citet{softbox} and \citet{gumbel_box} are improved training methods to handle the difficulties inherent in gradient-descent based region learning.
\citet{query2box} and \citet{BoxE} use a box-based adjustment of their loss functions, which suggest learning per-entity thresholds are beneficial. \citet{chen2021probabilistic} use box embeddings to model uncertain knowledge graphs,  \citet{onoe2021modeling} use boxes for fined grained entity typing, and \citet{patel2022modeling} use boxes for multi-label classification.

Fuzzy sets, a generalization of sets, have been widely studied in the context of clustering \cite{bezdek1978fuzzy}, decision theory \cite{zimmermann1987fuzzy} and linguistics \cite{de2000modelling}.  
However, the use of fuzzy sets in NLP has been fairly limited. \citet{reviewer_request} normalized each dimension of a word vector against all the word vectors in the vocabulary and interpret them as probability features that enabled them to perform fuzzy set theoretic operations with the words. \citet{zhao_fuzzy_2018} and \citet{zhelezniak_dont_2019} build fuzzy set representations of sentences using pre-trained vector embeddings for words and show the usefulness such representations on semantic textual similarity (STS) tasks.
\citet{jimenez_softcardinality-core_2013, jimenez_unal-nlp_2014} use the soft-cardinality features for a fuzzy set representation of a sentence to perform the task of entailment and textual relatedness. All these works use pre-trained vector embeddings for the words to form fuzzy sets representing sentences, which contrasts with this work where the emphasis is on learning fuzzy set representations for words from corpus. 
\section{Conclusion}
In this work we have demonstrated that box embeddings can not only effectively train to represent pairwise similarity but also can capture the rich set-theoretic structure of words via unsupervised training. This is a consequence of the fact that Gumbel boxes are an efficient parameterization of fuzzy sets, with sufficient representational capacity to model complicated co-occurrance interactions while, at the same time, allowing for tractable computation and gradient-based training of set-theoretic queries. The set-theoretic representation capabilities of box models allow them to generalize in a calibrated manner, leading to a more coherent and self-consistent model of sets.

\FloatBarrier

\section*{Acknowledgments}
The authors would like to thank the members of the Information and Extraction Synthesis Laboratory (IESL) at UMass Amherst for helpful discussions, as well as Katrin Erk and Ken Clarkson.
This work was partially supported by
IBM Research AI through the AI Horizons Network and the Chan Zuckerberg Initiative under the project Scientific Knowledge Base Construction. Additional support was provided by
the National Science Foundation (NSF) under Grant Numbers IIS-1514053 and IIS-2106391, 
the Defense Advanced Research Projects Agency (DARPA) via Contract No. FA8750-17-C-0106 under Subaward No. 89341790 from the University of Southern California, and the Office of Naval Research (ONR) via Contract No. N660011924032 under Subaward No. 123875727 from the University of Southern California.
The U.S. Government is authorized to reproduce and distribute reprints for Governmental purposes notwithstanding any copyright notation thereon. The views and conclusions contained herein are those of the authors and should not be interpreted as necessarily representing the official policies or endorsements, either expressed or implied, of IBM, CZI, NSF, DARPA, ONR, or the U.S. Government.


\bibliographystyle{bibliography/acl_natbib}
\bibliography{bibliography/custom,bibliography/fuzzy,bibliography/anthology,bibliography/zotero_references,bibliography/mendeley_references}

\begin{thebibliography}{53}
\expandafter\ifx\csname natexlab\endcsname\relax\def\natexlab#1{#1}\fi

\bibitem[{Abboud et~al.(2020)Abboud, Ceylan, Lukasiewicz, and Salvatori}]{BoxE}
Ralph Abboud, {\.I}smail~{\.I}lkan Ceylan, Thomas Lukasiewicz, and Tommaso
  Salvatori. 2020.
\newblock Boxe: A box embedding model for knowledge base completion.
\newblock In \emph{Proceedings of the 34th Annual Conference on Neural
  Information Processing Systems‚ NeurIPS}.

\bibitem[{Athiwaratkun and Wilson(2018)}]{gaussian_density}
Ben Athiwaratkun and Andrew~Gordon Wilson. 2018.
\newblock Hierarchical density order embeddings.
\newblock In \emph{International Conference on Learning Representations}.

\bibitem[{Baker et~al.(2014)Baker, Reichart, and Korhonen}]{verb-143}
Simon Baker, Roi Reichart, and Anna Korhonen. 2014.
\newblock \href {https://doi.org/10.3115/v1/D14-1034} {An unsupervised model
  for instance level subcategorization acquisition}.
\newblock In \emph{Proceedings of the 2014 Conference on Empirical Methods in
  Natural Language Processing ({EMNLP})}, pages 278--289, Doha, Qatar.
  Association for Computational Linguistics.

\bibitem[{Baroni et~al.(2012)Baroni, Bernardi, Do, and Shan}]{baroni-2012}
Marco Baroni, Raffaella Bernardi, Ngoc-Quynh Do, and Chung-chieh Shan. 2012.
\newblock \href {https://www.aclweb.org/anthology/E12-1004} {Entailment above
  the word level in distributional semantics}.
\newblock In \emph{Proceedings of the 13th Conference of the {E}uropean Chapter
  of the Association for Computational Linguistics}, pages 23--32, Avignon,
  France. Association for Computational Linguistics.

\bibitem[{Baroni et~al.(2009)Baroni, Bernardini, Ferraresi, and
  Zanchetta}]{wackypedia}
Marco Baroni, Silvia Bernardini, A.~Ferraresi, and E.~Zanchetta. 2009.
\newblock The wacky wide web: a collection of very large linguistically
  processed web-crawled corpora.
\newblock \emph{Language Resources and Evaluation}, 43:209--226.

\bibitem[{Bezdek and Harris(1978)}]{bezdek1978fuzzy}
James~C Bezdek and J~Douglas Harris. 1978.
\newblock Fuzzy partitions and relations; an axiomatic basis for clustering.
\newblock \emph{Fuzzy sets and systems}, 1(2):111--127.

\bibitem[{Bhat et~al.(2020)Bhat, Debnath, Banerjee, and
  Shrivastava}]{reviewer_request}
Siddharth Bhat, Alok Debnath, Souvik Banerjee, and Manish Shrivastava. 2020.
\newblock \href {https://doi.org/10.18653/v1/2020.repl4nlp-1.4} {Word
  embeddings as tuples of feature probabilities}.
\newblock In \emph{Proceedings of the 5th Workshop on Representation Learning
  for NLP}, pages 24--33, Online. Association for Computational Linguistics.

\bibitem[{Bojanowski et~al.(2017)Bojanowski, Grave, Joulin, and
  Mikolov}]{fasttext}
Piotr Bojanowski, Edouard Grave, Armand Joulin, and Tomas Mikolov. 2017.
\newblock Enriching word vectors with subword information.
\newblock \emph{Transactions of the Association for Computational Linguistics},
  5:135--146.

\bibitem[{Boratko et~al.(2021)Boratko, Burroni, Dasgupta, and
  McCallum}]{boratko2021min}
Michael Boratko, Javier Burroni, Shib~Sankar Dasgupta, and Andrew McCallum.
  2021.
\newblock Min/max stability and box distributions.
\newblock \emph{UAI}, pages 2146--2155.

\bibitem[{Brown et~al.(2020)Brown, Mann, Ryder, Subbiah, Kaplan, Dhariwal,
  Neelakantan, Shyam, Sastry, Askell et~al.}]{gpt3}
Tom Brown, Benjamin Mann, Nick Ryder, Melanie Subbiah, Jared~D Kaplan, Prafulla
  Dhariwal, Arvind Neelakantan, Pranav Shyam, Girish Sastry, Amanda Askell,
  et~al. 2020.
\newblock Language models are few-shot learners.
\newblock \emph{Advances in neural information processing systems},
  33:1877--1901.

\bibitem[{Bruni et~al.(2014)Bruni, Tran, and Baroni}]{men}
Elia Bruni, Nam~Khanh Tran, and Marco Baroni. 2014.
\newblock Multimodal distributional semantics.
\newblock \emph{J. Artif. Int. Res.}, 49(1):1–47.

\bibitem[{Chamberlain et~al.(2017)Chamberlain, Clough, and
  Deisenroth}]{hyperbolic_graph}
Benjamin~Paul Chamberlain, James~R. Clough, and Marc~Peter Deisenroth. 2017.
\newblock Neural embeddings of graphs in hyperbolic space.
\newblock \emph{ArXiv}, abs/1705.10359.

\bibitem[{Chen et~al.(2021)Chen, Boratko, Chen, Dasgupta, Li, and
  McCallum}]{chen2021probabilistic}
Xuelu Chen, Michael Boratko, Muhao Chen, Shib~Sankar Dasgupta, Xiang~Lorraine
  Li, and Andrew McCallum. 2021.
\newblock Probabilistic box embeddings for uncertain knowledge graph reasoning.
\newblock \emph{Proceedings of the 2019 Conference of the North {A}merican
  Chapter of the Association for Computational Linguistics}.

\bibitem[{Dasgupta et~al.(2020)Dasgupta, Boratko, Zhang, Vilnis, Li, and
  McCallum}]{gumbel_box}
Shib~Sankar Dasgupta, Michael Boratko, Dongxu Zhang, Luke Vilnis,
  Xiang~Lorraine Li, and Andrew McCallum. 2020.
\newblock Improving local identifiability in probabilistic box embeddings.
\newblock In \emph{Advances in Neural Information Processing Systems}.

\bibitem[{De~Cock et~al.(2000)De~Cock, Bodenhofer, and Kerre}]{de2000modelling}
Martine De~Cock, Ulrich Bodenhofer, and Etienne~E Kerre. 2000.
\newblock Modelling linguistic expressions using fuzzy relations.
\newblock In \emph{Proc. 6th Int. Conf. on Soft Computing (IIZUKA2000)}, pages
  353--360. Citeseer.

\bibitem[{Devlin et~al.(2019)Devlin, Chang, Lee, and Toutanova}]{bert}
Jacob Devlin, Ming-Wei Chang, Kenton Lee, and Kristina Toutanova. 2019.
\newblock {BERT}: Pre-training of deep bidirectional transformers for language
  understanding.
\newblock In \emph{Proceedings of the 2019 Conference of the North {A}merican
  Chapter of the Association for Computational Linguistics: Human Language
  Technologies, Volume 1 (Long and Short Papers)}, Minneapolis, Minnesota.
  Association for Computational Linguistics.

\bibitem[{Finkelstein et~al.(2001)Finkelstein, Gabrilovich, Matias, Rivlin,
  Solan, Wolfman, and Ruppin}]{ws353}
Lev Finkelstein, Evgeniy Gabrilovich, Yossi Matias, Ehud Rivlin, Zach Solan,
  Gadi Wolfman, and Eytan Ruppin. 2001.
\newblock Placing search in context: The concept revisited.
\newblock In \emph{Proceedings of the 10th international conference on World
  Wide Web}, pages 406--414.

\bibitem[{Ganea et~al.(2018)Ganea, B{\'e}cigneul, and
  Hofmann}]{hyperbolic_cone}
Octavian-Eugen Ganea, Gary B{\'e}cigneul, and Thomas Hofmann. 2018.
\newblock Hyperbolic entailment cones for learning hierarchical embeddings.
\newblock In \emph{International Conference on Machine Learning}.

\bibitem[{Halawi et~al.(2012)Halawi, Dror, Gabrilovich, and Koren}]{Mturk-771}
Guy Halawi, Gideon Dror, Evgeniy Gabrilovich, and Yehuda Koren. 2012.
\newblock Large-scale learning of word relatedness with constraints.
\newblock In \emph{Proceedings of the 18th ACM SIGKDD international conference
  on Knowledge discovery and data mining}, pages 1406--1414.

\bibitem[{Hill et~al.(2015)Hill, Reichart, and Korhonen}]{simlex999}
Felix Hill, Roi Reichart, and Anna Korhonen. 2015.
\newblock Simlex-999: Evaluating semantic models with (genuine) similarity
  estimation.
\newblock \emph{Computational Linguistics}, 41(4):665--695.

\bibitem[{Jimenez et~al.(2013)Jimenez, Becerra, and
  Gelbukh}]{jimenez_softcardinality-core_2013}
Sergio Jimenez, Claudia Becerra, and Alexander Gelbukh. 2013.
\newblock \href {https://aclanthology.org/S13-1028} {{SOFTCARDINALITY}-{CORE}:
  {Improving} {Text} {Overlap} with {Distributional} {Measures} for {Semantic}
  {Textual} {Similarity}}.
\newblock In \emph{Second {Joint} {Conference} on {Lexical} and {Computational}
  {Semantics} (*{SEM}), {Volume} 1: {Proceedings} of the {Main} {Conference}
  and the {Shared} {Task}: {Semantic} {Textual} {Similarity}}, pages 194--201,
  Atlanta, Georgia, USA. Association for Computational Linguistics.

\bibitem[{Jimenez et~al.(2014)Jimenez, Dueñas, Baquero, and
  Gelbukh}]{jimenez_unal-nlp_2014}
Sergio Jimenez, George Dueñas, Julia Baquero, and Alexander Gelbukh. 2014.
\newblock \href {https://doi.org/10.3115/v1/S14-2131} {{UNAL}-{NLP}:
  {Combining} {Soft} {Cardinality} {Features} for {Semantic} {Textual}
  {Similarity}, {Relatedness} and {Entailment}}.
\newblock In \emph{Proceedings of the 8th {International} {Workshop} on
  {Semantic} {Evaluation} ({SemEval} 2014)}, pages 732--742, Dublin, Ireland.
  Association for Computational Linguistics.

\bibitem[{Klir and Yuan(1996)}]{fuzzy-zadeh-1996}
George~J Klir and Bo~Yuan. 1996.
\newblock Fuzzy sets, fuzzy logic, and fuzzy systems: selected papers by lotfi
  a. zadeh.

\bibitem[{Lai and Hockenmaier(2017)}]{poe}
Alice Lai and Julia Hockenmaier. 2017.
\newblock Learning to predict denotational probabilities for modeling
  entailment.
\newblock In \emph{EACL}.

\bibitem[{Lee and Zadeh(1969)}]{note-on-fuzzy-language}
E.T. Lee and L.A. Zadeh. 1969.
\newblock \href {https://doi.org/https://doi.org/10.1016/0020-0255(69)90025-5}
  {Note on fuzzy languages}.
\newblock \emph{Information Sciences}, 1(4):421--434.

\bibitem[{Li et~al.(2019)Li, Vilnis, Zhang, Boratko, and McCallum}]{softbox}
Xiang Li, Luke Vilnis, Dongxu Zhang, Michael Boratko, and Andrew McCallum.
  2019.
\newblock Smoothing the geometry of probabilistic box embeddings.
\newblock \emph{ICLR}.

\bibitem[{Luong et~al.(2013)Luong, Socher, and Manning}]{rw}
Thang Luong, Richard Socher, and Christopher Manning. 2013.
\newblock \href {https://www.aclweb.org/anthology/W13-3512} {Better word
  representations with recursive neural networks for morphology}.
\newblock In \emph{Proceedings of the Seventeenth Conference on Computational
  Natural Language Learning}, pages 104--113, Sofia, Bulgaria. Association for
  Computational Linguistics.

\bibitem[{Meyer and Lewis(2020)}]{lexical-ambiguity}
Francois Meyer and Martha Lewis. 2020.
\newblock Modelling lexical ambiguity with density matrices.
\newblock \emph{arXiv preprint arXiv:2010.05670}.

\bibitem[{Mikolov et~al.(2013)Mikolov, Sutskever, Chen, Corrado, and
  Dean}]{word2vec}
Tomas Mikolov, Ilya Sutskever, Kai Chen, Greg~S Corrado, and Jeff Dean. 2013.
\newblock Distributed representations of words and phrases and their
  compositionality.
\newblock In \emph{NIPS}.

\bibitem[{Miller and Charles(1991)}]{MC}
George~A Miller and Walter~G Charles. 1991.
\newblock \href {http://eric.ed.gov/ERICWebPortal/recordDetail?accno=EJ431389}
  {Contextual correlates of semantic similarity}.
\newblock \emph{Language \& Cognitive Processes}, 6(1):1--28.

\bibitem[{Nelson et~al.(1980)Nelson, Mcevoy, Walling, and Wheeler}]{homographs}
Douglas Nelson, Cathy Mcevoy, John Walling, and Joseph Wheeler. 1980.
\newblock \href {https://doi.org/10.3758/BF03208320} {The university of south
  florida homograph norms}.
\newblock \emph{Behavior research methods, instruments, \& computers: a journal
  of the Psychonomic Society, Inc}, 12:16--37.

\bibitem[{Nickel and Kiela(2017)}]{poincare}
Maximilian Nickel and Douwe Kiela. 2017.
\newblock Poincar{\'e} embeddings for learning hierarchical representations.
\newblock In \emph{Neural Information Processing Systems}.

\bibitem[{Onoe et~al.(2021)Onoe, Boratko, and Durrett}]{onoe2021modeling}
Yasumasa Onoe, Michael Boratko, and Greg Durrett. 2021.
\newblock Modeling fine-grained entity types with box embeddings.
\newblock \emph{Association for Computational Linguistics}.

\bibitem[{Patel et~al.(2022)Patel, Dangati, Lee, Boratko, and
  McCallum}]{patel2022modeling}
Dhruvesh Patel, Pavitra Dangati, Jay-Yoon Lee, Michael Boratko, and Andrew
  McCallum. 2022.
\newblock \href {https://openreview.net/forum?id=tyTH9kOxcvh} {Modeling label
  space interactions in multi-label classification using box embeddings}.
\newblock In \emph{International Conference on Learning Representations}.

\bibitem[{Pennington et~al.(2014)Pennington, Socher, and Manning}]{glove}
Jeffrey Pennington, Richard Socher, and Christopher Manning. 2014.
\newblock {G}lo{V}e: Global vectors for word representation.
\newblock In \emph{Proceedings of the 2014 Conference on Empirical Methods in
  Natural Language Processing ({EMNLP})}. Association for Computational
  Linguistics.

\bibitem[{Peters et~al.(2018)Peters, Neumann, Iyyer, Gardner, Clark, Lee, and
  Zettlemoyer}]{elmo}
Matthew~E. Peters, Mark Neumann, Mohit Iyyer, Matt Gardner, Christopher Clark,
  Kenton Lee, and Luke Zettlemoyer. 2018.
\newblock \href {http://arxiv.org/abs/1802.05365} {Deep contextualized word
  representations}.
\newblock \emph{CoRR}, abs/1802.05365.

\bibitem[{Radford et~al.(2019)Radford, Wu, Child, Luan, Amodei, Sutskever
  et~al.}]{gpt2}
Alec Radford, Jeffrey Wu, Rewon Child, David Luan, Dario Amodei, Ilya
  Sutskever, et~al. 2019.
\newblock Language models are unsupervised multitask learners.
\newblock \emph{OpenAI blog}, 1(8):9.

\bibitem[{Radinsky et~al.(2011)Radinsky, Agichtein, Gabrilovich, and
  Markovitch}]{Mturk-287}
Kira Radinsky, Eugene Agichtein, Evgeniy Gabrilovich, and Shaul Markovitch.
  2011.
\newblock A word at a time: computing word relatedness using temporal semantic
  analysis.
\newblock In \emph{Proceedings of the 20th international conference on World
  wide web}, pages 337--346.

\bibitem[{Ren et~al.(2020)Ren, Hu, and Leskovec}]{query2box}
Hongyu Ren, Weihua Hu, and Jure Leskovec. 2020.
\newblock Query2box: Reasoning over knowledge graphs in vector space using box
  embeddings.
\newblock In \emph{8th International Conference on Learning Representations}.
  OpenReview.net.

\bibitem[{Ren and Leskovec(2020)}]{ren2020beta}
Hongyu Ren and Jure Leskovec. 2020.
\newblock Beta embeddings for multi-hop logical reasoning in knowledge graphs.
\newblock \emph{arXiv preprint arXiv:2010.11465}.

\bibitem[{Rubenstein and Goodenough(1965)}]{rg}
Herbert Rubenstein and John~B. Goodenough. 1965.
\newblock \href {https://doi.org/10.1145/365628.365657} {Contextual correlates
  of synonymy}.
\newblock \emph{Commun. ACM}, 8(10):627–633.

\bibitem[{Suzuki et~al.(2019)Suzuki, Takahama, and
  Onoda}]{suzuki2019hyperbolic}
Ryota Suzuki, Ryusuke Takahama, and Shun Onoda. 2019.
\newblock Hyperbolic disk embeddings for directed acyclic graphs.
\newblock In \emph{International Conference on Machine Learning}, pages
  6066--6075. PMLR.

\bibitem[{Tifrea et~al.(2019)Tifrea, Becigneul, and Ganea}]{poincare_glove}
Alexandru Tifrea, Gary Becigneul, and Octavian-Eugen Ganea. 2019.
\newblock \href {https://openreview.net/forum?id=Ske5r3AqK7} {Poincare glove:
  Hyperbolic word embeddings}.
\newblock In \emph{International Conference on Learning Representations}.

\bibitem[{Vendrov et~al.(2016)Vendrov, Kiros, Fidler, and
  Urtasun}]{order_embedding}
Ivan Vendrov, Ryan Kiros, Sanja Fidler, and Raquel Urtasun. 2016.
\newblock Order-embeddings of images and language.
\newblock In \emph{International Conference on Learning Representations}.

\bibitem[{Vilnis et~al.(2018)Vilnis, Li, Murty, and McCallum}]{hard_box}
Luke Vilnis, Xiang Li, Shikhar Murty, and Andrew McCallum. 2018.
\newblock Probabilistic embedding of knowledge graphs with box lattice
  measures.
\newblock In \emph{Association for Computational Linguistics}.

\bibitem[{Vilnis and McCallum(2015)}]{vilnis2015word}
Luke Vilnis and Andrew McCallum. 2015.
\newblock Word representations via gaussian embedding.
\newblock In \emph{ICLR}.

\bibitem[{{Wikipedia contributors}(2022)}]{homographs_wikipedia}
{Wikipedia contributors}. 2022.
\newblock List of english homographs --- {Wikipedia}{,} the free encyclopedia.
\newblock
  \url{https://en.wikipedia.org/w/index.php?title=List_of_English_homographs&oldid=1074954944}.
\newblock [Online; accessed 9-March-2022].

\bibitem[{Yang and Powers(2006)}]{YP}
Dongqiang Yang and David M.~W. Powers. 2006.
\newblock Verb similarity on the taxonomy of wordnet.
\newblock In \emph{In the 3rd International WordNet Conference (GWC-06), Jeju
  Island, Korea}.

\bibitem[{Zadeh(1965)}]{zadeh1965fuzzy}
L.A. Zadeh. 1965.
\newblock \href {https://doi.org/https://doi.org/10.1016/S0019-9958(65)90241-X}
  {Fuzzy sets}.
\newblock \emph{Information and Control}, 8(3):338--353.

\bibitem[{Zhao and Mao(2018)}]{zhao_fuzzy_2018}
Rui Zhao and Kezhi Mao. 2018.
\newblock \href {https://doi.org/10.1109/TFUZZ.2017.2690222} {Fuzzy
  {Bag}-of-{Words} {Model} for {Document} {Representation}}.
\newblock \emph{IEEE Transactions on Fuzzy Systems}, 26(2):794--804.
\newblock Conference Name: IEEE Transactions on Fuzzy Systems.

\bibitem[{Zhelezniak et~al.(2019{\natexlab{a}})Zhelezniak, Savkov, Shen,
  Moramarco, Flann, and Hammerla}]{zhelezniak2018dont}
Vitalii Zhelezniak, Aleksandar Savkov, April Shen, Francesco Moramarco, Jack
  Flann, and Nils~Y. Hammerla. 2019{\natexlab{a}}.
\newblock \href {https://openreview.net/forum?id=SkxXg2C5FX} {Don't settle for
  average, go for the max: Fuzzy sets and max-pooled word vectors}.
\newblock In \emph{International Conference on Learning Representations}.

\bibitem[{Zhelezniak et~al.(2019{\natexlab{b}})Zhelezniak, Savkov, Shen,
  Moramarco, Flann, and Hammerla}]{zhelezniak_dont_2019}
Vitalii Zhelezniak, Aleksandar Savkov, April Shen, Francesco Moramarco, Jack
  Flann, and Nils~Y Hammerla. 2019{\natexlab{b}}.
\newblock \href {https://openreview.net/forum?id=SkxXg2C5FX} {{DON}’{T}
  {SETTLE} {FOR} {AVERAGE}, {GO} {FOR} {THE} {MAX}: {FUZZY} {SETS} {AND}
  {MAX}-{POOLED} {WORD} {VECTORS}}.
\newblock In \emph{International {Conference} on {Learning} {Representations}}.

\bibitem[{Zimmermann(1987)}]{zimmermann1987fuzzy}
Hans-J{\"u}rgen Zimmermann. 1987.
\newblock \emph{Fuzzy sets, decision making, and expert systems}, volume~10.
\newblock Springer Science \& Business Media.

\end{thebibliography}

\appendix
\section{Preprocessing}
\label{app: preprocessing}
The WaCKypedia corpus has been tokenized and lemmatized. We used the lemmatized version of the corpus, however it was observed that various tokens were not split as they should have been (eg. "1.5billion" -> "1.5 billion"). We split tokens using regex criteria to identify words and numbers. All punctuation was removed from the corpus, all numbers were replaced with a "<num>" token, and all words were made lowercase. We also removed any words which included non-ascii symbols. After this step, the entire corpus was tokenized once more, and any token occurring less than 100 times was dropped.

\section{Dataset Analysis}
\label{app:dataset_analysis}
\begin{table}[ht!]
\begin{tabular}{|l|l|}
\hline
Dataset      & \begin{tabular}[c]{@{}l@{}}Median\\ \\ Frequency\end{tabular} \\ \hline
Men-Tr-3K    & 23942                                                         \\ \hline
Mc-30        & 25216                                                      \\ \hline
Mturk-771    & 43128                                                       \\ \hline
Simlex-999   & 40653                                                       \\ \hline
Verb-143     & 309192                                                      \\ \hline
Yp-130       & 23044                                                       \\ \hline
Rw-Stanford  & 5683                                                        \\ \hline
Rg-65        & 13088                                                       \\ \hline
Ws-353-All   & 58803                                                       \\ \hline
Ws-353-Sim.  & 57514                                                       \\ \hline
Ws-353-Rel   & 64490                                                      \\ \hline
Mturk-287    & 32952                                                         \\ \hline
Simverb-3500 & 39020                                                         \\ \hline
\end{tabular}
\caption{Median Frequency of each similarity dataset.}
\end{table}

\section{Hyperparameters}
\label{app:training_hyperparam}
As discussed in Section \ref{sec:exp-results}, we train on 128 dimensional \wv and 64 dimensional \wb models for 10 epochs. We ran at least 60 runs for each of the models with random seed and randomly chose hyperparamter from the following range -
batch\_size:[2048, 4096, 8192, 16384, 32768], learning rate  log\_uniform[exp(-1), exp(-10)], 
Window\_size: [5, 6, 7, 8, 9, 10], negative\_samples: [2, 5, 10, 20], sub\_sampling threshold: [0.001, 0.0001]. The best working hyperparameter sets and the corresponding checkpoints can be found here: 

\section{Dataset for Set Theoretic Queries}
\label{app:set_theoretic_queries}
In this section, we describe the dataset for set theoretic evaluation. We evaluate on \textit{set-intersection}, \textit{set-difference}, \textit{set-union} queries. For each of these tasks, we create queries of the form $<A, B, A \circ B>$, where, $\circ$ is any of the mentioned set operation. In case of \textit{set-union}, we find homographs to be an excellent choice as they are words describing multiple different choices of words. We choose commonly used homographs from list of homographs available in wikipedia \cite{homographs_wikipedia} to construct this dataset. We manually eliminated many of the words which are rare or when the homographs are referring to concepts which are semantically similar. We provide some examples of the dataset for union queries in table \ref{tab:union}. Also, note that we can perform the task for \textit{set-difference} by just swapping the $B$ and $A \cup B$, since $B = (A \cup B) \setminus A$, i.e., if we subtract one concept from the homographs then we must get back a set containing the other concept. So the same table is being used for \textit{set-difference} task. We manually create a small evaluation set for the \textit{set-intersection} task, listed in Table \ref{tab:intersection}.
\begin{table}
\begin{tabular}{|l|l|l|}
\hline
$A$                   & $B$                      & $A \cap B$                  \\ \hline
\textit{girl}       & \textit{boy}           & \textit{child}       \\
\textit{pet}        & \textit{wolf}          & \textit{dog}         \\
\textit{winner}     & \textit{medal}         & \textit{gold}        \\
\textit{video}      & \textit{entertainment} & \textit{movie}       \\
\textit{ocean}      & \textit{sound}         & \textit{wave}        \\
\textit{finance}    & \textit{river}         & \textit{bank}        \\
\textit{parent}     & \textit{woman}         & \textit{mother}      \\
\textit{bird}       & \textit{America}       & \textit{eagle}       \\
\textit{car}        & \textit{sea}           & \textit{boat}        \\
\textit{farm}       & \textit{animal}        & \textit{cow}         \\
\textit{fruit}      & \textit{yellow}        & \textit{banana}      \\
\textit{house}      & \textit{royal}         & \textit{palace}      \\
\textit{property}   & \textit{chemistry}     & \textit{solubility}  \\
\textit{bank}       & \textit{river}         & \textit{basin}       \\
\textit{policy}     & \textit{government}    & \textit{legislation} \\
\textit{incense}    & \textit{odor}          & \textit{candle}      \\
\textit{spirit}     & \textit{drink}         & \textit{beer}        \\
\textit{dance}      & \textit{song}          & \textit{ballad}      \\
\textit{work}       & \textit{art}           & \textit{painting}    \\
\textit{instrument} & \textit{wind}          & \textit{flute}       \\ \hline
\end{tabular}
\caption{Set theoretic queries for \textit{Intersection}. In this task, given $A$ and $B$, the model need to predict the word for $A \cap B$.}
\label{tab:intersection}
\end{table}

\begin{table}[!ht]
\begin{tabular}{|l|l|l|}
\hline
$A$               & $B$                & $A\cup B$               \\ \hline
\textit{table}    & \textit{chair}     & \textit{furniture}      \\
\textit{car}      & \textit{plane}     & \textit{transportation} \\
\textit{city}     & \textit{village}   & \textit{location}       \\
\textit{wolf}     & \textit{bear}      & \textit{animal}         \\
\textit{shirt}    & \textit{pant}      & \textit{clothes}        \\
\textit{computer} & \textit{phone}     & \textit{electronics}    \\
\textit{red}      & \textit{blue}      & \textit{color}          \\
\textit{movie}    & \textit{book}      & \textit{entertainment}  \\
\textit{school}   & \textit{college}   & \textit{education}      \\
\textit{doctor}   & \textit{engineer}  & \textit{profession}     \\
\textit{box}      & \textit{circle}    & \textit{shape}          \\
\textit{big}      & \textit{small}     & \textit{size}           \\
\textit{dog}      & \textit{tree}      & \textit{bark}           \\
\textit{fish}     & \textit{tone}      & \textit{bass}           \\
\textit{sports}   & \textit{wing}      & \textit{bat}            \\
\textit{carry}    & \textit{animal}    & \textit{bear}           \\
\textit{sadness}  & \textit{color}     & \textit{blue}           \\
\textit{bend}     & \textit{weapon}    & \textit{bow}            \\
\textit{hit}      & \textit{food}      & \textit{buffet}         \\
\textit{combine}  & \textit{building}  & \textit{compound}       \\
\textit{happy}    & \textit{list}      & \textit{content}        \\
\textit{acquire}  & \textit{agreement} & \textit{contract}       \\ \hline
\end{tabular}
\caption{Examples of set theoretic queries for \textit{Union}. In this task, given $A$ and $B$, the model need to predict the word for $A \cap B$. Also note that, we use the same table for the \textit{set difference} queries by treating swapping the $B$ and $A \cup B$ columns.}
\label{tab:union}
\end{table}

\end{document}